\documentclass[journal]{IEEEtai}

\usepackage[colorlinks,urlcolor=blue,linkcolor=blue,citecolor=blue]{hyperref}

\usepackage{color,array}
\usepackage{graphicx}

\usepackage{cite}
\usepackage{amsmath,amssymb,amsfonts}
\usepackage{algorithm}
\usepackage{algorithmic}
\usepackage{textcomp}
\usepackage{xcolor}
\def\BibTeX{{\rm B\kern-.05em{\sc i\kern-.025em b}\kern-.08em
    T\kern-.1667em\lower.7ex\hbox{E}\kern-.125emX}}
\usepackage{booktabs}
\usepackage{multirow}
\usepackage[super]{nth}
\usepackage{tikz}
\newcommand*\circled[1]{\tikz[baseline=(char.base)]{
		\node[shape=circle,draw,inner sep=0.2pt] (char) {#1};}}
\newcommand*\circledB[1]{\tikz[baseline=(char.base)]{
            \node[shape=circle,fill,inner sep=0.2pt] (char) {\textcolor{white}{#1}};}}
\usetikzlibrary{tikzmark}
\tikzset{circledColor/.style={circle,draw,inner sep=0.1em,line width=0.04em}}
\let\labelindent\relax
\usepackage{soul}
\usepackage{enumitem}

\usepackage{setspace}
\usepackage{cleveref}
 
\begin{document}
 
\title{SpikeNAS: A Fast Memory-Aware Neural Architecture Search Framework for Spiking Neural Network-based Embedded AI Systems} 

\author{Rachmad Vidya Wicaksana Putra, \IEEEmembership{Member, IEEE}, and Muhammad Shafique, \IEEEmembership{Senior Member, IEEE}
\thanks{Rachmad Vidya Wicaksana Putra is with eBrain Lab, Division of Engineering, New York University (NYU) Abu Dhabi, United Arab Emirates;
{e-mail: rachmad.putra@nyu.edu}
}%
\thanks{Muhammad Shafique is the Director of eBrain Lab, Division of Engineering, New York University (NYU) Abu Dhabi, United Arab Emirates;
{e-mail: muhammad.shafique@nyu.edu}
}
\vspace{-0.6cm}
}

\markboth{To Appear at the IEEE Transactions on Artificial Intelligence (TAI) 2025}
{First A. Author \MakeLowercase{\textit{et al.}}: Bare Demo of IEEEtai.cls for IEEE Journals of IEEE Transactions on Artificial Intelligence}

\maketitle


\begin{abstract}
Embedded AI systems are expected to incur low power/energy consumption for solving machine learning tasks, as these systems are usually power constrained (e.g., object recognition task in autonomous mobile agents with portable batteries).
These requirements can be fulfilled by Spiking Neural Networks (SNNs), since their bio-inspired spike-based operations offer high accuracy and ultra low-power/energy computation. 
Currently, most of SNN architectures are derived from Artificial Neural Networks whose neurons' architectures and operations are different from SNNs, and/or developed without considering memory budgets from the underlying processing hardware of embedded platforms. 
These limitations hinder SNNs from reaching their full potential in accuracy and efficiency. 
Toward this, we propose \textit{SpikeNAS}, a novel fast memory-aware neural architecture search (NAS) framework for SNNs that quickly finds an appropriate SNN architecture with high accuracy under the given memory budgets from targeted embedded systems.
To do this, our SpikeNAS employs several key steps: analyzing the impacts of network operations on the accuracy, enhancing the network architecture to improve the learning quality, developing a fast memory-aware search algorithm, and performing quantization. 
The experimental results show that our SpikeNAS improves the searching time and maintains high accuracy compared to state-of-the-art while meeting the given memory budgets (e.g., 29x, 117x, and 3.7x faster search for CIFAR10, CIFAR100, and TinyImageNet200 respectively, using an Nvidia RTX A6000 GPU machine), thereby quickly providing the appropriate SNN architecture for the memory-constrained embedded AI systems.
\end{abstract}

\begin{IEEEImpStatement}
Developing SNN models for the memory- and power-constrained embedded AI systems is highly desirable due to its immense benefits, but very challenging to do. 
However, manual SNN development is laborious and time-consuming, while state-of-the-art NAS techniques for SNNs do not consider the memory budget from the targeted hardware platform.
To address this, our NAS technique quickly generates an SNN that meets memory budget and offers high accuracy by identifying the impact of network operations, enhancing the network architecture, and developing a fast search algorithm. 
In this manner, our research contributes to the development of application-specific SNNs under any given hardware platforms, thereby enabling their applicability for diverse application use-cases.
\end{IEEEImpStatement}


\section{Introduction}

\begin{figure}[t]
\centering
\includegraphics[width=\linewidth]{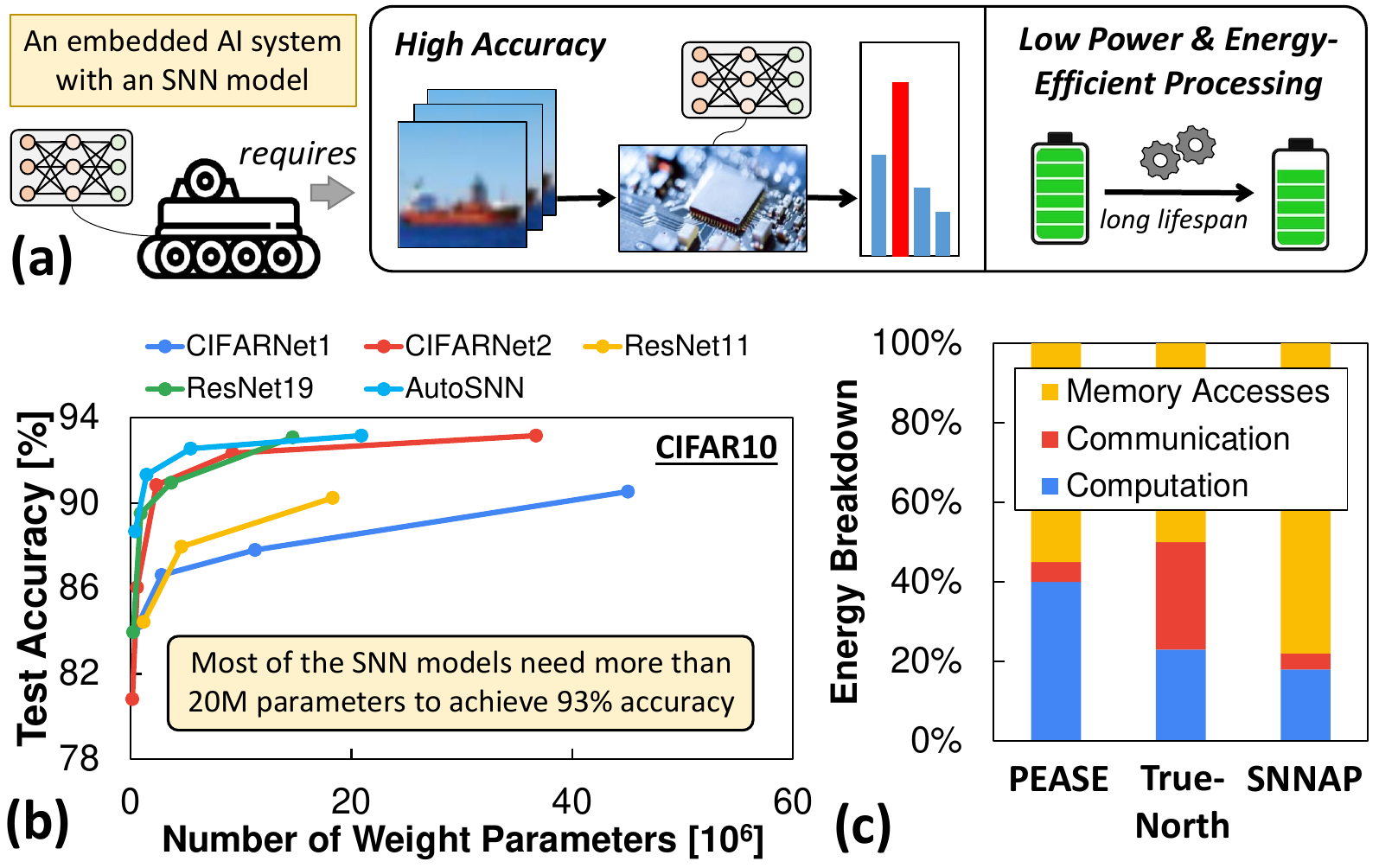}
\vspace{-0.7cm}
\caption{(a) Embedded AI systems are expected to achieve high accuracy with low power and energy-efficient processing to complete ML-based tasks. 
(b) Accuracy and memory of different SNN models for CIFAR10: CIFARNet1~\cite{Ref_Wu_DirectTrainSNNs_AAAI19}, CIFARNet2~\cite{Ref_Fang_MemTConstantSNNs_ICCV21}, ResNet11~\cite{Ref_Lee_SpikeBackprop_FNINS20}, ResNet19~\cite{Ref_Zheng_LargerSNNs_AAAI21}, AutoSNN~\cite{Ref_Na_AutoSNN_ICML22}; adapted from studies in~\cite{Ref_Na_AutoSNN_ICML22}. 
(c) Breakdown of energy consumption for SNN processing by different hardware platforms: PEASE~\cite{Ref_Roy_PEASE_ISLPED17}, TrueNorth~\cite{Ref_Akopyan_TrueNorth_TCAD15}, and SNNAP~\cite{Ref_Sen_ApproxSNN_DATE17}; adapted from studies in~\cite{Ref_Krithivasan_SpikeBundle_ISLPED19}.}
\label{Fig_SNN_AccMem_HWenergy}
\vspace{-0.7cm}
\end{figure}

\IEEEPARstart{E}{mdedded} systems are expected to incorporate neural network (NN) algorithms~\cite{Ref_Roy_SpikeMachineIntel_Nature19, Ref_Schuman_NeuroCompute_Nature22, Ref_Bartolozzi_EmbodiedNeuroIntel_Nature22, Ref_Putra_NeuromorphicAIRobotics_arXiv24}, since these algorithms can achieve state-of-art accuracy in solving diverse machine learning (ML) tasks, such as image classification~\cite{Ref_Putra_SpikeDyn_DAC21}, object recognition~\cite{Ref_Cordone_ObjDetSNN_IJCNN22}, and continual learning~\cite{Ref_Putra_lpSpikeCon_IJCNN22}; referred to as the \textit{embedded artificial intelligence (AI) systems}.
Therefore, these systems need to incur low power and energy-efficient consumption when executing NN algorithms, because they are usually powered by portable batteries~\cite{Ref_Schuman_NeuroCompute_Nature22}\cite{Ref_Bartolozzi_EmbodiedNeuroIntel_Nature22}.
For instance, battery-powered autonomous mobile agents (e.g., UGVs) need to achieve high accuracy with  efficient NN processing to preserve their battery lifetime; see Fig.~\ref{Fig_SNN_AccMem_HWenergy}(a).  
These requirements can be fulfilled by Spiking Neural Networks (SNNs), as their bio-inspired spike-based operations offer high accuracy while incurring ultra low-power/energy computation~\cite{Ref_Putra_FSpiNN_TCAD20}. 
Therefore, many research works are carried out to maximize the SNN potentials, ranging from SNN architecture developments to its deployments for diverse ML-based applications~\cite{Ref_Na_AutoSNN_ICML22}\cite{Ref_Davies_Loihi_IEEEMM18, Ref_Lin_ProgrammingLoihi_IEEEMC18, Ref_Putra_QSpiNN_IJCNN21, Ref_Shafique_EdgeAI_ICCAD21, Ref_Putra_ReSpawn_ICCAD21, Ref_Putra_SparkXD_DAC21, Ref_Safa_SNN4Radar_TNNLS21, Ref_Safa_NeuromorphicNearSensor_Micro22, Ref_Putra_EnforceSNN_FNINS22, Ref_Basu_SNNicSurvey_CICC22, Ref_Kim_SNASNet_ECCV22, Ref_Zhang_AUVwithSNN_FNBOT22, Ref_Putra_RescueSNN_FNINS23, Ref_Putra_TopSpark_IROS23, Ref_Rathi_SNNsurvey_CSUR23, Ref_Moitra_SpikeSim_TCAD23, Ref_Putra_SNN4Agents_FROBT24, Ref_Bano_SNNParametersStudy_arXiv24, Ref_Bano_FastSpiker_arXiv24, Ref_Ma_Darwin3_NSC24}. 
In fact, different ML-based applications usually have different accuracy requirements, which often necessitate different memory costs to facilitate the corresponding SNN architectures. 
For instance, a bigger SNN is typically developed to achieve a higher accuracy, while incurring a larger memory footprint, as shown in Fig.~\ref{Fig_SNN_AccMem_HWenergy}(b). 
Furthermore, a larger memory footprint leads to a higher energy consumption, as the memory accesses still dominate the overall processing energy of SNNs; see Fig.~\ref{Fig_SNN_AccMem_HWenergy}(c). 
Therefore, \textit{the development of SNN architectures that can achieve the expected accuracy with acceptable memory footprint is needed to support the widespread use of SNNs for embedded AI systems}. 

Currently, most of the SNN architectures are derived from Artificial Neural Networks (ANNs), whose neurons' architectures and operations are different from SNNs~\cite{Ref_Na_AutoSNN_ICML22}. 
SNNs accumulate and generate temporal information in form of spikes using Leaky Integrate-and-Fire (LIF) neurons, while ANNs perform multiplication-and-accumulation (MAC) and activation using Rectified Linear Unit (ReLU).
Such ANN-based SNN architectures may lose unique features from SNN computation models (e.g., temporal information between spikes), thereby leading to sub-optimal accuracy~\cite{Ref_Kim_SNASNet_ECCV22}.  
Moreover, the SNN architectures are still developed without considering memory budgets from the underlying processing hardware, which is important to ensure the applicability of the SNN architecture for the given hardware platform. 
For instance, most of the SNN models require more than 2x10$^7$ (20M) of parameters to achieve 93\% accuracy for the CIFAR10 dataset, as shown in Fig.~\ref{Fig_SNN_AccMem_HWenergy}(b), which is memory consuming and inefficient for resource-constrained embedded systems. 
\textit{These limitations hinder the SNNs from reaching their full potential in accuracy and efficiency for embedded AI systems}.
However, manually designing the desired SNN architecture is a laborious task and time consuming, thus requiring an alternative solution that quickly provides an SNN architecture that is suitable for the given accuracy and memory constraints.

\textit{\textbf{Research Problem:} 
How to automatically and quickly find an SNN architecture that achieves acceptable accuracy under the given memory constraints for embedded AI systems}.
An efficient solution to this problem will ease SNN developments for embedded AI systems, thus enabling the widespread use of SNNs for diverse ML-based application use-cases~\cite{Ref_Putra_NeuromorphicAIRobotics_arXiv24}.

\begin{figure}[t]
\centering
\includegraphics[width=0.92\linewidth]{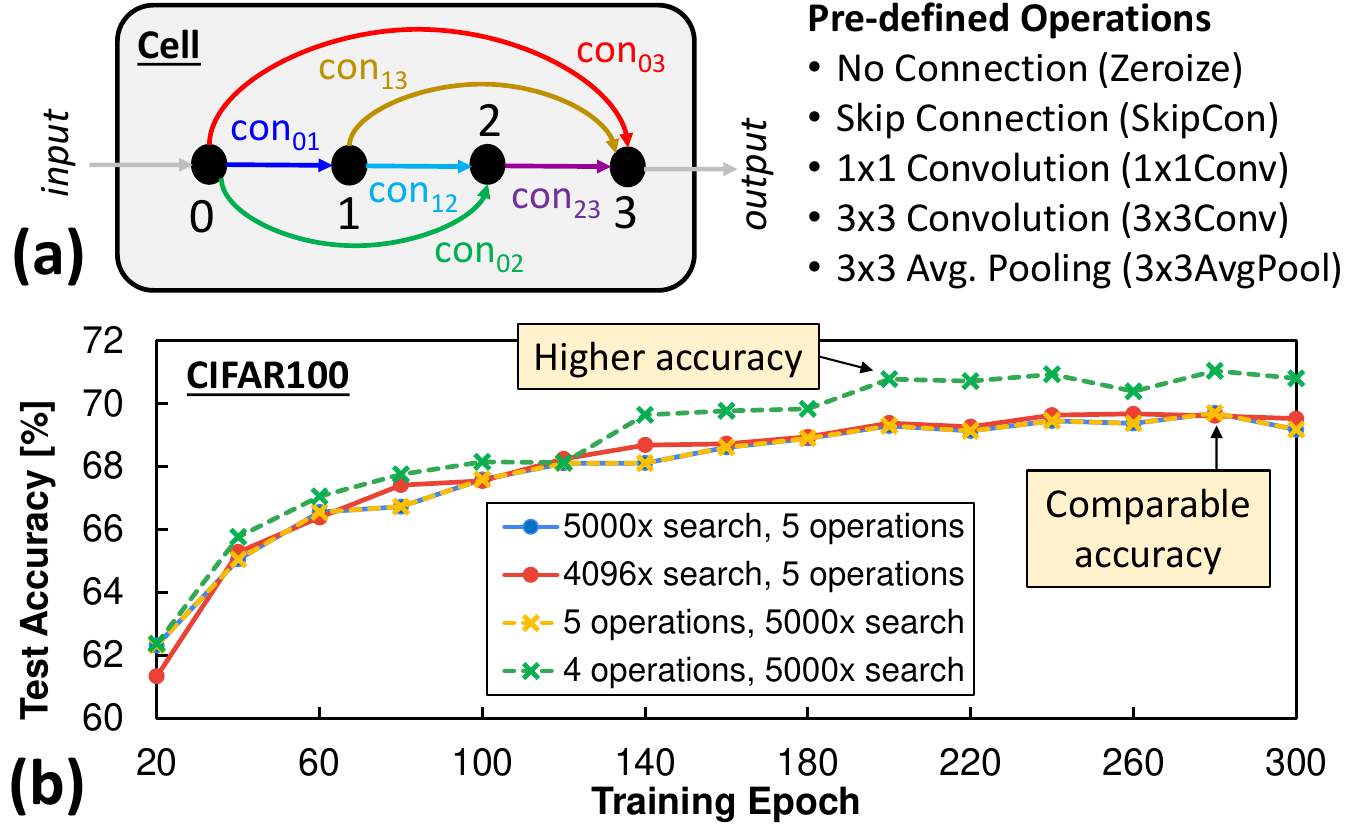}
\vspace{-0.3cm}
\caption{(a) A cell is a directed acyclic graph whose edge corresponds to an operation from the pre-defined operations~\cite{Ref_Dong_NASbench201_ICLR20}.
(b) Results of accuracy across different numbers of search iteration and different operation types.}
\label{Fig_Cell_ObserveNsrchNops}
\vspace{-0.5cm}
\end{figure}
\vspace{-0.2cm}
\subsection{State-of-the-Art and Their Limitations}
\label{Sec_Intro_SOA}

To address the research problem, previous works employ neural architecture search (NAS) to automatically find SNN architectures that can achieve high accuracy. 
Currently, NAS for SNNs is still at the early stage, hence only a few studies found in the literature, including~\cite{Ref_Na_AutoSNN_ICML22}\cite{Ref_Kim_SNASNet_ECCV22}\cite{Ref_Pan_EvoNAS4SNN_arxiv23, Ref_Che_AutoSpikeformer_arXiv23, Ref_Putra_SNNSizeScaling_arXiv24}. 
The prominent one is SNASNet~\cite{Ref_Kim_SNASNet_ECCV22}, which finds an SNN architecture that achieves high accuracy by employing a NAS without training approach to reduce the searching time. 
However, \textit{these state-of-the-art works do not consider memory constraints from the target hardware in their search algorithms, thereby limiting the efficiency of the generated SNN architectures when deployed on the target hardware}. 
To identify and better understand the limitations of state-of-the-art (i.e., the NAS without training approach in~\cite{Ref_Kim_SNASNet_ECCV22}), we perform an experimental case study and discuss it in Section~\ref{Sec_Intro_CaseStudy}.

\vspace{-0.2cm}
\subsection{Case Study and Research Challenges}
\label{Sec_Intro_CaseStudy}

We study the impacts of search iterations and operation types on the accuracy considering a NAS without training approach using the search space and mechanism of SNASNet, i.e, exploring the topology and operations of a cell architecture in Fig.~\ref{Fig_Cell_ObserveNsrchNops}(a) using a random search. 
To evaluate the impact of search iterations, the state-of-the-art work employs 5000x iterations and the investigated one employs 4096x. 
To evaluate the impact of operation types, the state-of-the-art work employs 5 operation types in Fig.~\ref{Fig_Cell_ObserveNsrchNops}(a), while the investigated one employs 4 operation types (i.e., without Zeroize).  
Details of the experimental setup are provided in Section~\ref{Sec_Evaluation}.
The experimental results are presented in Fig.~\ref{Fig_Cell_ObserveNsrchNops}(b), from which we draw the following key observations.
\begin{itemize}[leftmargin=*]
    \item A smaller number of search iterations may lead to comparable accuracy across different training epochs, meaning that there is a possibility to maintain accuracy with a smaller search space.
    \item A smaller number of operation types may lead to comparable (or better) accuracy across different training epochs, meaning that there is a possibility to maintain (or improve) accuracy with a smaller search space.
\end{itemize}

These observations expose several \textit{research challenges and requirements (outlined below)} in devising solutions for the targeted research problem.
\begin{itemize}[leftmargin=*]
    \item The solution should quickly perform the search to minimize the searching time.
    \item The solution should be able to guide the search mechanism towards finding an SNN architecture that maintains high accuracy, while avoiding redundancy.
    \item The solution should incorporate the memory constraints into its searching algorithm to ensure the applicability and efficiency of the SNN deployments on the targeted hardware.
\end{itemize}

\vspace{-0.3cm}
\subsection{Our Novel Contributions}
\label{Sec_Intro_NovelContrib}

\begin{figure}[t]
\centering
\includegraphics[width=\linewidth]{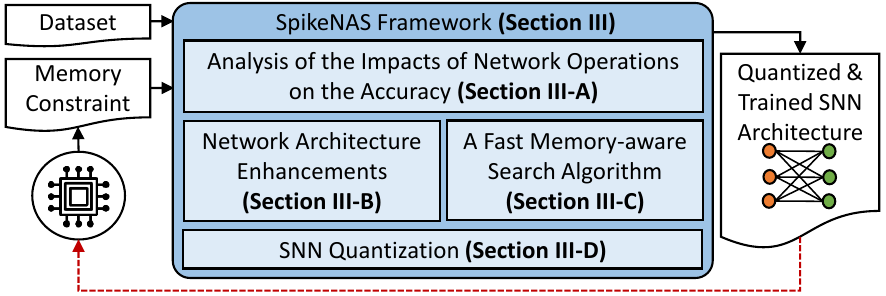}
\vspace{-0.7cm}
\caption{Overview of our novel contributions (in blue boxes).}
\label{Fig_SpikeNAS_Novelty}
\vspace{-0.5cm}
\end{figure}

\begin{figure*}[t]
\centering
\includegraphics[width=\linewidth]{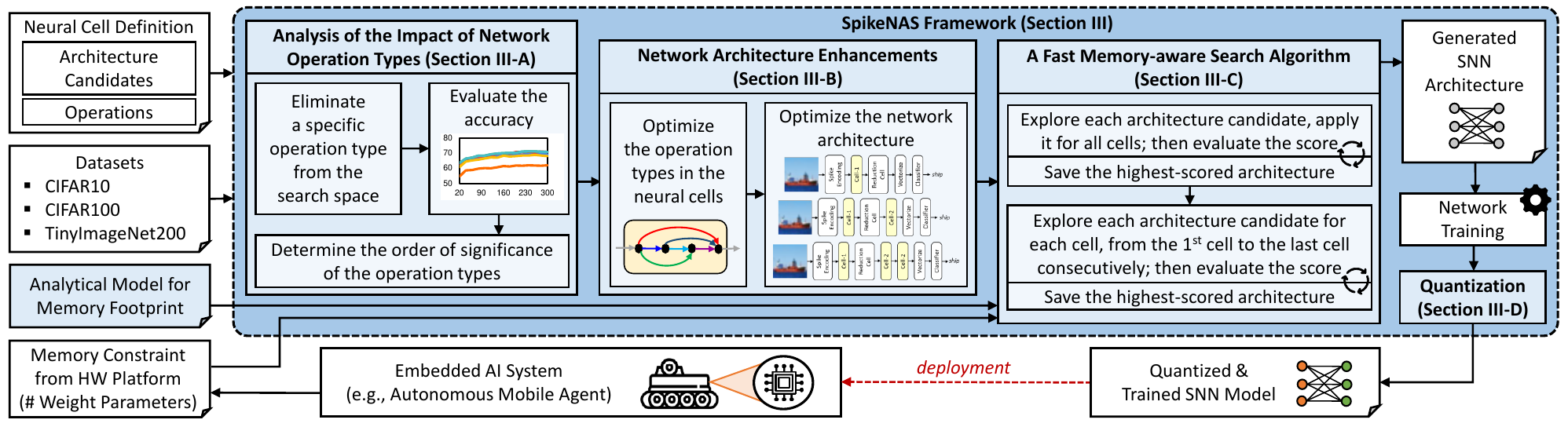}
\vspace{-0.7cm}
\caption{Overview of our SpikeNAS framework, and its key steps for generating an SNN architecture that offers high accuracy while meeting the given memory constraint from an embedded AI system (e.g., autonomous mobile agent). The novel contributions are highlighted in blue boxes.}
\label{Fig_SpikeNAS_Framework}
\vspace{-0.4cm}
\end{figure*}

Towards this, we propose \textit{SpikeNAS}, a novel NAS framework that can quickly find the appropriate SNN architecture with high accuracy under the given memory budgets from the hardware platforms. 
It employs several key steps as the following (an overview is in Fig.~\ref{Fig_SpikeNAS_Novelty} and the details in Fig.~\ref{Fig_SpikeNAS_Framework}). 
\begin{itemize}[leftmargin=*]
    \item \textbf{Analysis of the Impacts of Network Operations (Section~\ref{Sec_SpikeNAS_Analysis}):} 
    It aims to understand the significance of each pre-defined operation type, i.e., by removing the investigated operation type from the search space and then observing the output accuracy.
    \item \textbf{Network Architecture Enhancements (Section~\ref{Sec_SpikeNAS_Enhance}):} 
    It aims to maintain or improve accuracy of the network architecture, while reducing the searching time.  
    It is done by optimizing the cell operation types and the number of cells in the network architectures.
    \item \textbf{A Fast Memory-aware Search Algorithm (Section~\ref{Sec_SpikeNAS_Search}):} 
    It aims to find an SNN architecture that achieves high accuracy and meets the memory budget, by performing search for each individual cell, while monitoring the memory cost of the investigated architecture.
    \item \textbf{SNN Quantization (Section~\ref{Sec_SpikeNAS_Quant}):}
    It further reduces the memory footprint of the generated SNN model through the post-training quantization (PTQ) technique, while considering the accepted accuracy degradation.
\end{itemize}  

\textbf{Key Results:} 
We evaluate our SpikeNAS framework using a Python-based implementation on an  Nvidia RTX A6000 GPU machine. 
The experimental results show that our SpikeNAS improves the searching time and maintains high accuracy compared to state-of-the-art while meeting the memory budgets, e.g., our SpikeNAS improves the searching time by 29x, 117x, and 3.7x for CIFAR10, CIFAR100, and TinyImageNet200, respectively.

\vspace{-0.2cm}
\section{Preliminaries}
\label{Sec_Prelim}

\subsection{Spiking Neural Networks (SNNs)}
\label{Sec_Prelim_SNNs}

SNNs are considered the third generation of NN computation model~\cite{Ref_Putra_FSpiNN_TCAD20}. 
SNNs typically employ a LIF neuron model due to their bio-plausible data representation and computation~\cite{Ref_Putra_ReSpawn_ICCAD21}\cite{Ref_Putra_RescueSNN_FNINS23}\cite{Ref_Tavanaei_DLSNN_Neunet18, Ref_Pfeiffer_DLSNN_FNINS18, Ref_Putra_SoftSNN_DAC22, Ref_Putra_SpiKernel_ESL24}, whose neuronal dynamics can be formulated as follows.

\vspace{-0.1cm}
\begin{equation}
    \begin{split}
    \tau_{leak} \frac{dV_{mem}(t)}{dt} &  = - (V_{mem}(t)-V_{reset}) + x(t) 
    \end{split}
    \label{Eq_LIF}
\end{equation}
\begin{equation}
    \begin{split}
    \text{if} \;\;\; V_{mem} & \geq V_{th} \;\;\; \text{then} \;\;\; V_{mem} \leftarrow V_{reset}
    \end{split}
    \label{Eq_LIF_fire}
\end{equation}
$V_{mem}$ denotes the neurons' membrane potential at timestep-$t$, $V_{reset}$ denotes the neurons' reset potential, and $\tau_{leak}$ denotes the time constant of membrane potential decay. 
Meanwhile, $x$ denotes the inputs, which can be stated in a discrete time form as the following.

\begin{equation}
    \begin{split}
    x^l[t] = \sum_i wgh_i^l \; \delta_i^{l-1}[t]+b^l
    \end{split}
    \label{Eq_LIF_input}
\end{equation}
$x$ denotes the neuron input for layer-$l$ at timestep-$t$, $b$ denotes the bias, and $wgh$ denotes the weight at index-$i$.
Meanwhile, $\delta$ denotes the input spike from previous layer ($l-1$) at index-$i$ and timestep-$t$. 

\vspace{-0.2cm}
\subsection{Neural Architecture Search (NAS) for SNNs}
\label{Sec_Prelim_NAS}

NAS techniques for SNNs~\cite{Ref_Na_AutoSNN_ICML22}\cite{Ref_Kim_SNASNet_ECCV22}\cite{Ref_Pan_EvoNAS4SNN_arxiv23, Ref_Che_AutoSpikeformer_arXiv23, Ref_Putra_SNNSizeScaling_arXiv24}\cite{Ref_Wang_AutoST_ICASSP24} can be classified into two approaches, i.e., ``\textit{NAS with training}'' and ``\textit{NAS without training}''. 

\smallskip
\textbf{NAS with Training:}
In this category, the prominent work is the AutoSNN~\cite{Ref_Na_AutoSNN_ICML22}, which aims at finding an SNN architecture with high accuracy and low number of spikes. 
It trains a super-network that includes all of the possible candidates of SNN architectures, then performs a search to find a subset of network architecture that achieves the target. 
However, this NAS approach requires a huge searching time as well as a huge memory and energy consumption, as it needs to perform training for a large network before searching and determining the desired SNN architecture.

\smallskip
\textbf{NAS without Training:}
Initially, NAS without training approach was proposed in ANN domain~\cite{Ref_Mellor_NASnoTraining_ICML21}, and then leveraged for SNN domain~\cite{Ref_Kim_SNASNet_ECCV22}.
In this category, the prominent work is the SNASNet~\cite{Ref_Kim_SNASNet_ECCV22}, whose key idea is to quickly measure the capability of a network to reach high accuracy at early stage, thereby avoiding the training phase to evaluate the network. 
It is motivated by the observation that an architecture with distinctive representations across different samples is likely to achieve high accuracy after training~\cite{Ref_Mellor_NASnoTraining_ICML21}. 
To measure the representation capability of a network, the SNASNet employs the following key steps (our SpikeNAS framework also follows the same steps for evaluating the network).
\begin{itemize}[leftmargin=*]
    \item It employs LIF neuron to divide the input space into active and inactive regions. If $V_{mem}$ reaches $V_{th}$ then the neuron is mapped to 1, otherwise 0. 
    Then, the activity of LIF neurons in each layer is represented as binary vector $f$. 
    \item It feeds $S$ samples in a mini-batch to the network. 
    Then, a kernel matrix $\textbf{K}_H$ for each layer is constructed by computing the Hamming distance $H(f_i, f_j)$ between different samples $i$ and $j$; see Eq.~\ref{Eq_Kmatrix}.
    Here, $F$ is the number of LIF neurons in the given layer, while $\alpha$ is the factor to normalize high sparsity from LIF neurons.
    \item It calculates the score of the architecture candidate using Eq.~\ref{Eq_Kscore}. 
    Then, the candidate with the highest score is selected for training.
\end{itemize} 
\vspace{0.1cm}
\begin{equation}
 \textbf{K}_H = 
 \begin{pmatrix}
   F-\alpha H(f_1,f_1) & \cdots & F - \alpha H(f_1,f_S) \\
   \vdots   & \ddots & \vdots  \\
   F - \alpha H(f_S,f_1)  & \cdots & F - \alpha H(f_S,f_S) 
 \end{pmatrix}
 \label{Eq_Kmatrix}
\end{equation} 
\begin{equation}
  \begin{split}
    score = \log (\det \; \lvert \sum_l \textbf{K}_H^l  \rvert)
    \end{split}
    \label{Eq_Kscore}
\end{equation}
To achieve this, SNASNet explores the architecture candidates of the \textit{cell} (i.e., a directed acyclic graph that represents network topology and operations); see Fig.~\ref{Fig_Cell_ObserveNsrchNops}(a). 
Afterward, it evaluates the score of the investigated architecture, then trains the architecture with the highest score~\cite{Ref_Kim_SNASNet_ECCV22}.
However, it has several limitations as follows.
\begin{itemize}[leftmargin=*]
    \item It employs \textit{a random search} mechanism to determine the cell architecture, hence the effectiveness and duration of the search depend on the number of search iteration. 
    \item It has a fixed number of cells (i.e., 2 cells) as shown in Fig.~\ref{Fig_SNASNet_NetArch} and it applies the same architecture for all cells, thus reducing the possibility to find a better SNN architecture (e.g., higher accuracy).
\end{itemize}

\vspace{-0.3cm}
\subsection{Neural Cell-based Strategy for NAS}
\label{Sec_Prelim_Cell}

\begin{figure}[t]
\centering
\includegraphics[width=0.95\linewidth]{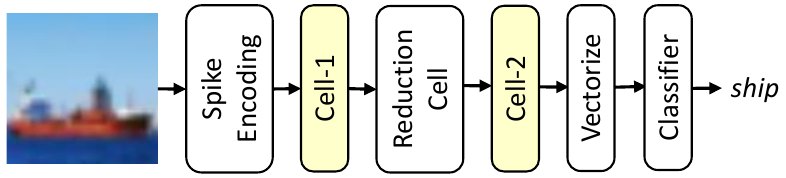}
\vspace{-0.3cm}
\caption{The macro-architecture for a cell-based NAS in the state-of-the-art (i.e., employing 2 cells).}
\label{Fig_SNASNet_NetArch}
\vspace{-0.6cm}
\end{figure}

\textit{Neural cell} (or simply \textit{cell})-based strategy has been widely used for NAS in ANNs. 
The motivation of cell-based NAS is to provide a unified benchmark for NAS algorithms~\cite{Ref_Dong_NASbench201_ICLR20}. 
Its approach is to search for the connection topology and the corresponding operations inside each cell; see Fig.~\ref{Fig_Cell_ObserveNsrchNops}(a). 
A cell typically consists of 4 nodes and 5 pre-defined operation types: No-Connection (Zeroize), Skip-Connection (SkipCon), 1x1 Convolution (1x1Conv), 3x3 Convolution (3x3Conv), and 3x3 Average Pooling (3x3AvgPool). 
The connection between 2 nodes employs a specific pre-defined operation type. 
Therefore, there are 15625 architecture candidates in each cell. 
The state-of-the-art work (i.e., SNASNet) employs a 2-cell SNN architecture for its NAS technique; see its macro-architecture in Fig.~\ref{Fig_SNASNet_NetArch}. 
In this work, we also employ the same macro-architecture for our SpikeNAS framework, and focus on the feedforward topology as it has been widely used in the SNN community.

\vspace{-0.2cm}
\section{SpikeNAS Framework}
\label{Sec_SpikeNAS}

To address the above-discussed research problem and challenges, we propose a novel SpikeNAS framework; see the detailed flow in Fig.~\ref{Fig_SpikeNAS_Framework}. 
Its key steps are discussed in the Section~\ref{Sec_SpikeNAS_Analysis} - Section~\ref{Sec_SpikeNAS_Search}, consecutively.

\vspace{-0.4cm}
\subsection{Analysis of the Impact of Operation Types}
\label{Sec_SpikeNAS_Analysis}

\begin{figure}[t]
\centering
\includegraphics[width=\linewidth]{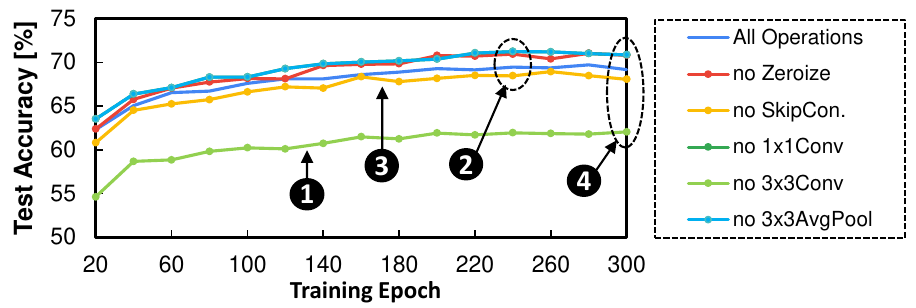}
\vspace{-0.7cm}
\caption{Results of removing different types of operations from the search space of a cell architecture. These results are for CIFAR100, and results for CIFAR10 show similar trends.}
\label{Fig_SpikeNAS_ObserveOps}
\vspace{-0.5cm}
\end{figure}

Discussion in Section~\ref{Sec_Intro_CaseStudy} suggests that employing all pre-defined operations might not lead to optimal solution. 
Therefore, to understand the significance of each pre-defined operation, we perform an experimental case study. 
Here, we remove a specific operation type, then perform NAS considering a macro-architecture in Fig.~\ref{Fig_SNASNet_NetArch} with 5000x search iteration. 
Since we employ the NAS without training approach, then this step is performed before the search process and the training phase.
Experimental results are shown in Fig.~\ref{Fig_SpikeNAS_ObserveOps}, from which we draw the following key observations. 

\begin{itemize}
    \item[\circledB{1}] Removing the 3x3Conv operation from the cells significantly reduces the accuracy, as this operation employs 3x3 kernel size for extracting important features from input feature maps. 
    Hence, removing the 3x3Conv make the network difficult to perform learning properly. 
    Therefore, \textit{the 3x3Conv operation should be kept in the search space}.
    \item[\circledB{2}] Removing the Zeroize, 1x1Conv, 3x3AvgPool operations from the cells do not change the accuracy significantly. 
    Therefore, \textit{these operations may be removed from the search space}.
    \item[\circledB{3}] Removing the SkipCon operation from the search space also notably reduces the accuracy for CIFAR100, as this operation provides feature maps from previous layer to ensure the preservation of important information. 
    This observation also aligns with the observation from studies in~\cite{Ref_Benmeziane_SkipInSNNs_ScaDL23}.
    Therefore, \textit{the SkipCon operation should be kept in the search space}.
    \item[\circledB{4}] In general, the order of significance of the operations from the highest to the lowest: (1) 3x3Conv, (2) SkipCon, and (3) Zeroize, 1x1Conv, and 3x3AvgPool.  
    We consider Zeroize, 1x1Conv, and 3x3AvgPool in the same significance level as they have comparable impact on the accuracy.
    \textit{These operations may be removed to reduce the search space based on their significance.} 
\end{itemize}

Afterward, these observations are then leveraged as guidance for enhancing the NAS, so that it can quickly find and generate SNN architecture that has significant operations or characteristics to achieve high accuracy, while considering the given memory constraints. 

\vspace{-0.2cm}
\subsection{Network Architecture Enhancements}
\label{Sec_SpikeNAS_Enhance}

\subsubsection{Optimization of the Operation Types in the Neural Cells}
\label{Sec_SpikeNAS_Enhance_Cell}

\begin{figure}[t]
\centering
\includegraphics[width=0.7\linewidth]{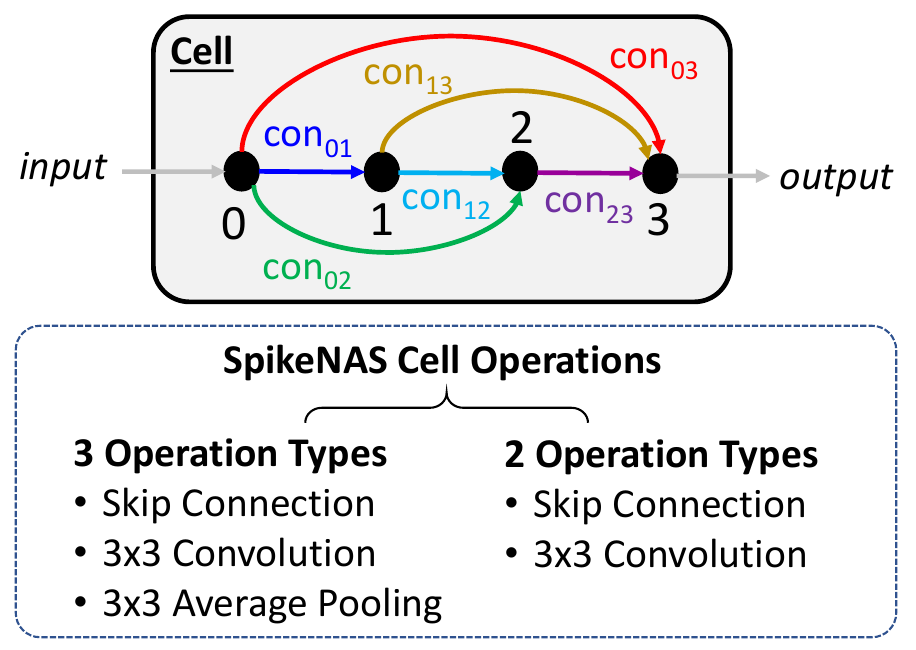}
\vspace{-0.3cm}
\caption{The pre-defined cell operation types in SpikeNAS. 
First operation set consists of 3 operation types, while second operation set consists of 2 operation types.}
\label{Fig_SpikeNAS_CellArch}
\vspace{-0.5cm}
\end{figure}

In Section~\ref{Sec_SpikeNAS_Analysis}, we observe that some operations can be removed from the search space to improve the searching time without sacrificing accuracy. 
Hence, \textit{we optimize the number of operation types based on their significance on the accuracy}.
We propose two variants of operation sets that can be considered in the search space to provide multiple optimization options (see an overview in Fig.~\ref{Fig_SpikeNAS_CellArch}), as follows. 
\begin{itemize}[leftmargin=*]
    \item \textbf{First operation set:} SkipCon, 3x3Conv, and 3x3AvgPool. We select 3x3AvgPool to reduce the size of feature maps, while minimizing the compute and memory requirements and ensuring the connectivity to the next layer.  
    \item \textbf{Second operation set:} SkipCon and 3x3Conv. 
\end{itemize}

\subsubsection{SNN Macro-Architecture Adjustments}
\label{Sec_SpikeNAS_Enhance_Net}

\begin{figure}[hbtp]
\vspace{-0.2cm}
\centering
\includegraphics[width=\linewidth]{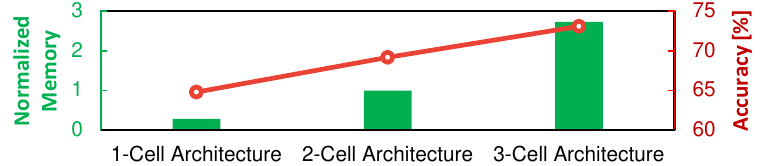}
\vspace{-0.7cm}
\caption{The accuracy vs memory trade-off among the 1-cell, 2-cell, and 3-cell architectures.}
\label{Fig_SpikeNAS_CellNumber}
\vspace{-0.6cm}
\end{figure}

\begin{figure}[t]
\centering
\includegraphics[width=\linewidth]{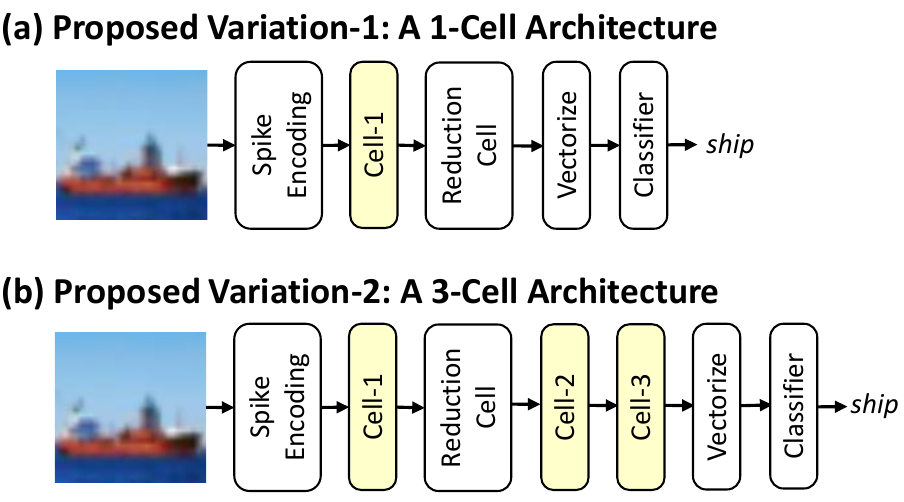}
\vspace{-0.7cm}
\caption{The proposed variations of the network architectures in SpikeNAS with (a) 1 cell and (b) 3 cells.}
\label{Fig_SpikeNAS_NetArch}
\vspace{-0.4cm}
\end{figure}

To increase the applicability and efficiency of SNN deployments for diverse ML-based applications, we propose adjustments for the SNN macro-architecture. 
Apart from the 2-cell SNN architecture, we also propose different variants of network architectures based on the number of cells (i.e., 1 cell and 3 cells), as discussed in the following.
\begin{itemize}[leftmargin=*]
    \item \textbf{1-Cell Architecture:} 
    It aims at reducing the memory footprint, but might reduce the accuracy as compared to the 2-cell architecture; see an illustration in Fig.~\ref{Fig_SpikeNAS_CellNumber}. 
    Its key idea is removing one cell from the 2-cell architecture; see an overview in Fig.~\ref{Fig_SpikeNAS_NetArch}(a).
    \item \textbf{3-Cell Architecture:} 
    It aims at improving the accuracy, but might incur a larger memory footprint as compared to the 2-cell architecture; see an illustration in Fig.~\ref{Fig_SpikeNAS_CellNumber}.
    Its key idea is adding one cell to the 2-cell architecture;
    see an overview in Fig.~\ref{Fig_SpikeNAS_NetArch}(b). 
\end{itemize}

In this manner, \textit{our SpikeNAS framework provides multiple design options to trade-off accuracy, memory footprint, and searching time}. 

\subsection{A Fast Memory-aware Search Algorithm}
\label{Sec_SpikeNAS_Search}

\textit{To synergistically perform architecture search while considering the memory constraint, we propose a fast memory-aware search algorithm.} 
The key ideas of our search algorithm are discussed in the following, while the corresponding pseudo-code is shown in Alg.~\ref{Alg_MemSpikingNAS}.
\begin{itemize}[leftmargin=*]
    \item We employ a dynamic programming approach to perform architecture search. 
    Specifically, we search and determine the architecture for each cell one-by-one (see Alg.~\ref{Alg_MemSpikingNAS}: line~2). 
    Therefore, more network candidates from different combinations of architectures across multiple cells are explored, hence leading to a higher possibility to find an SNN architecture that achieves high accuracy.
    \item Investigation of redundant architectures for each cell is minimized to improve the effectiveness of our architecture search. 
    To enable this, we propose an \textit{in-cell architecture search} (as shown in Alg.~\ref{Alg_MemSpikingNAS}: line 3).
    Here, variables $con_{xy}$ with $xy$ $\in \{01, 02, 03, 12, 13, 23\}$ correspond to the type of operation that connect node-$X$ to node-$Y$, as shown in Fig.~\ref{Fig_SpikeNAS_CellArch}. 
    Then, the cell architecture is constructed once $con_{xy}$ connections are defined (see Alg.~\ref{Alg_MemSpikingNAS}: line 4).
    \item All cells are initiated with the same architecture (see Alg.~\ref{Alg_MemSpikingNAS}: line 5-6). 
    Then, the architecture candidates are explored for each cell, starting from the second cell to the last one, consecutively (see Alg.~\ref{Alg_MemSpikingNAS}: line 7-8).
    Hence, the search space is reduced while keeping good architectures from previous cells.  
    Then, the network architecture is constructed once all cells are defined (Alg.~\ref{Alg_MemSpikingNAS}: line 10), and its memory cost is also evaluated (Alg.~\ref{Alg_MemSpikingNAS}: line 11).
    \item Afterward, the memory-constraining process is performed by incorporating the maximum allowed number of parameters into the search (see Alg.~\ref{Alg_MemSpikingNAS}: line 12).
    It is used to select which architectures should be evaluated, hence filtering out the ones that are not suitable and optimizing the searching time even further. 
    If the number of parameters from the investigated network ($mem_{cost}$) is less than or equal to the memory constraint/budget ($mem_{budget}$), then the network evaluation and selection are performed (see Alg.~\ref{Alg_MemSpikingNAS}: line 13-16).
    To do this, \textit{an analytical model of memory footprint is developed and employed into the search}. Details of the proposed analytical model is discussed in the following. 
\end{itemize}

\smallskip
\textbf{Analytical Model for Memory Footprint:}
We focus on the number of parameters in the SNN architecture ($N_{param}$) to represent the memory footprint, which can be calculated using Eq.~\ref{Eq_Nparam_all}-\ref{Eq_Nwgh_layer}.
\begin{equation}
  \begin{split}
    N_{param} = \sum_l (N_{wgh}^l+N_b^l)
    \end{split}
    \label{Eq_Nparam_all}
  \vspace{-0.3cm}
\end{equation}
\begin{equation}
  \begin{split}
    N_{wgh} = h_{wgh} \cdot w_{wgh} \cdot c_{wgh} \cdot r_{wgh}
    \end{split}
    \label{Eq_Nwgh_layer}
\end{equation}

$N_{wgh}^l$ and $N_b^l$ denotes the number of weights and bias in layer-$l$, respectively. 
The 2-dimensional size of filters in the given layer is represented with $h_{wgh}$ for height and $w_{wgh}$ for width.
Meanwhile, $c_{wgh}$ and $r_{wgh}$ denote the number of channel and the number of filters in the given layer, respectively. 
Furthermore, the memory footprint can also be represented in bit or byte format if required ($M_{param}$), by considering the bit precision ($bit$) as stated in Eq.~\ref{Eq_Mparam_all}.
\begin{equation}
  \begin{split}
    M_{param} = N_{param} \cdot bit
    \end{split}
    \label{Eq_Mparam_all}
    \vspace{-1cm}
\end{equation}

\begin{algorithm}[t]
  \footnotesize
  \caption{Our Fast Memory-aware Search Algorithm}
  \label{Alg_MemSpikingNAS}
  \begin{algorithmic}[1]
    \renewcommand{\algorithmicrequire}{\textbf{INPUT:}}
    \renewcommand{\algorithmicensure}{\textbf{OUTPUT:}}
	\REQUIRE 
        \textbf{(1)} Pre-defined number of neural cells ($C$); \\ 
        \textbf{(2)} Neural cells: object names ($cell_1$, ..., $cell_C$), cell architecture ($arch$); \\
        \textbf{(3)} Memory constraint or budget ($mem_{budget}$); \\
        \textbf{(4)} Pre-defined number of operation types ($T$); \\
        \textbf{(5)} Operation ($P$): SkipCon=0, 3x3Conv=1, and 3x3AvgPool=2; hence $P \in \{0, 1, 2\}$ \\
	\ENSURE Generated SNN architecture ($net\_arch$); \\
	\textbf{BEGIN} \\
	  \textbf{Initialization}: \\
	  \STATE $score_h$=-1000; \\ 
	  \textbf{Process}: \\
	  \FOR{($i$=1; $i$$<$($C$+1); $i$++)}
            \FOR{each combination from ($\forall$ $con_{01}$ $\in$ $P$), 
             ($\forall$ $con_{02}$ $\in$ $P$), 
             ($\forall$ $con_{03}$ $\in$ $P$), 
             ($\forall$ $con_{12}$ $\in$ $P$),
             ($\forall$ $con_{13}$ $\in$ $P$), and
             ($\forall$ $con_{23}$ $\in$ $P$)}
                \STATE $arch$ = Cell($con_{01}, con_{02}, con_{03}, con_{12}, con_{13}, con_{23}$); \\
                // create the cell architecture
		    \IF{($i$==1)}
		        \STATE $cell_1, ..., cell_N \leftarrow arch$; // apply $arch$ to all cells\\ 
		    \ELSE
		        \STATE $cell_i \leftarrow arch$; // apply $arch$ to $cell_i$\\
		    \ENDIF\\
                \STATE $net$ = Net$(cell_1, ..., cell_C)$; // create the network
                \STATE $mem_{cost}$= calc\_mem($net$); // calculate the memory cost
                \IF{($mem_{cost} \leq mem_{budget}$)}
                    \STATE $score =$ eval$(net)$; // evaluate the network score
                    \IF{($score > score_h$)}
                        \STATE $score_h = score$;\\
                        \STATE $net\_arch = net$;\\
                    \ENDIF\\
                \ENDIF\\
            \ENDFOR\\
	  \ENDFOR\\
	\RETURN $net\_arch$; \\
	\textbf{END}
	\end{algorithmic} 
\end{algorithm}
\setlength{\textfloatsep}{6pt}

\vspace{-0.6cm}
\subsection{SNN Quantization}
\label{Sec_SpikeNAS_Quant}
\vspace{-0.4cm}

To further reduce the memory footprint of generated SNNs, we quantize the SNN weights through a post-training quantization (PTQ) approach.  
To realize the quantization, we can employ the desired number format and rounding scheme. 
For instance, we consider the fixed-point format with truncation scheme, as they offer high accuracy while enabling fast design space exploration for SNN processing~\cite{Ref_Putra_EnforceSNN_FNINS22}.
The fixed-point format can be represented in $b$-bit precision and follows the 2's complement method, while the truncation (\textit{TR}) scheme for the given number $x$ under $b$-bit precision can be represented as \textit{TR}($x$,$b$) = $\lfloor x \rfloor$~\cite{Ref_Putra_QSpiNN_IJCNN21}.
However, quantizing the weight value also implies a reduction of its representation capability, thereby potentially decreasing the accuracy~\cite{Ref_Putra_FSpiNN_TCAD20}. 
To address this, we consider the trade-off between accuracy and memory footprint to determine the acceptable quantization levels (i.e., bit precision). 
Therefore, we can select the acceptable accuracy and memory footprint to meet with the design specifications.

The above discussion shows that, the SpikeNAS framework aims to enable a fast memory-aware NAS for generating a suitable SNN based on the memory constraint (i.e., maximum allowed number of parameters) and the NAS without training approach, which has not been comprehensively studied by state-of-the-art works. 
To achieve this, our SpikeNAS performs new studies and employs novel optimization techniques that are tailored specifically for SNNs, including analysis of the impact of SNN operations, network architecture enhancements, and a fast memory-aware search algorithm. 
Moreover, we also enhanced the SpikeNAS framework with post-training quantization (PTQ) to further reduce the memory footprint of the generated SNN. 
In this manner, our SpikeNAS framework integrates new studies, novel optimization techniques, and a widely-used PTQ technique to achieve high accuracy and high energy efficiency SNN processing.

\vspace{-0.2cm}
\section{Evaluation Methodology}
\label{Sec_Evaluation}

\begin{figure}[hbtp]
\vspace{-0.2cm}
\centering
\includegraphics[width=\linewidth]{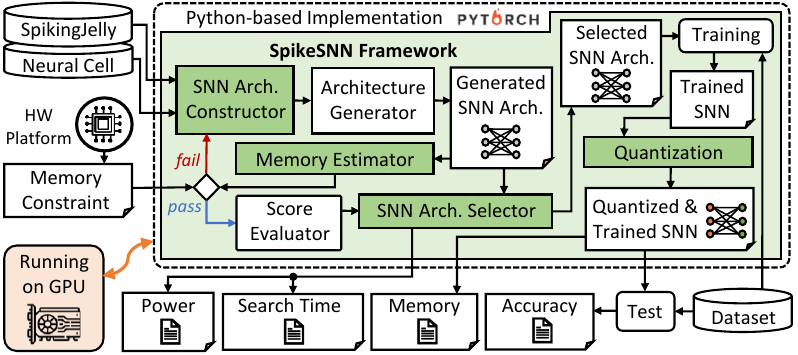}
\vspace{-0.7cm}
\caption{The experimental setup used for the evaluation.}
\label{Fig_Eval_Method}
\vspace{-0.4cm}
\end{figure}

To evaluate our SpikeNAS framework, we deploy the experimental setup as presented in Fig.~\ref{Fig_Eval_Method}, while considering the same evaluation conditions as widely used in the SNN community (e.g., focusing on the feedforward topology)~\cite{Ref_Nunes_SNNsurvey_Access22}. 
We evaluate different scenarios for SpikeNAS as presented in Table~\ref{Tab_EvalScenario1} and \ref{Tab_EvalScenario2} to provide a comprehensive study.
For brevity, we use an encoding for these scenarios: ``SpikeNAS\_\textit{p}C\textit{q}O'' refers to an architecture with \textit{p} cell(s) and \textit{q} operation types without memory constraint, while ``SpikeNAS\_\textit{p}C\textit{q}O\_M'' refers to the one with memory constraint.
We consider 1.2x10$^6$ (1.2M), 2x10$^6$ (2M), and 4x10$^6$ (4M) parameters as the memory constraints for CIFAR10, CIFAR100, and TinyImageNet200, respectively.
For the comparison partner, we consider SNASNet~\cite{Ref_Kim_SNASNet_ECCV22} and its default settings for the search space: 2 cells, 5 pre-defined operation types, and 5000x search iterations using a random search for experiments (i.e., SNASNet\_2C5O).  

\begin{table}[hbtp]
\caption{Different scenarios for SpikeNAS with 3 operation types: SkipCon, 3x3Conv, and 3x3AvgPool.}
\vspace{-0.1cm}
\label{Tab_EvalScenario1}
\footnotesize
\begin{tabular}{|ccc|}
\hline
\multicolumn{3}{|c|}{\textbf{Number of Cells}} \\ \hline
\multicolumn{1}{|c|}{\textbf{1}} & \multicolumn{1}{c|}{\textbf{2}} & \textbf{3} \\ \hline
\multicolumn{1}{|c|}{\begin{tabular}[c]{@{}c@{}}SpikeNAS\_1C3O\\ SpikeNAS\_1C3O\_M\end{tabular}} & \multicolumn{1}{c|}{\begin{tabular}[c]{@{}c@{}}SpikeNAS\_2C3O\\ SpikeNAS\_2C3O\_M\end{tabular}} & \begin{tabular}[c]{@{}c@{}}SpikeNAS\_3C3O\\ SpikeNAS\_3C3O\_M\end{tabular} \\ \hline
\end{tabular}
\vspace{-0.2cm}
\end{table}

\begin{table}[hbtp]
\vspace{-0.2cm}
\caption{Different scenarios for SpikeNAS with 2 operation types: SkipCon and 3x3Conv.}
\vspace{-0.1cm}
\label{Tab_EvalScenario2}
\footnotesize
\begin{tabular}{|ccc|}
\hline
\multicolumn{3}{|c|}{\textbf{Number of Cells}} \\ \hline
\multicolumn{1}{|c|}{\textbf{1}} & \multicolumn{1}{c|}{\textbf{2}} & \textbf{3} \\ \hline
\multicolumn{1}{|c|}{\begin{tabular}[c]{@{}c@{}}SpikeNAS\_1C2O\\ SpikeNAS\_1C2O\_M\end{tabular}} & \multicolumn{1}{c|}{\begin{tabular}[c]{@{}c@{}}SpikeNAS\_2C2O\\ SpikeNAS\_2C2O\_M\end{tabular}} & \begin{tabular}[c]{@{}c@{}}SpikeNAS\_3C2O\\ SpikeNAS\_3C2O\_M\end{tabular} \\ \hline
\end{tabular}
\vspace{-0.1cm}
\end{table}

\begin{figure*}[t]
\centering
\includegraphics[width=\linewidth]{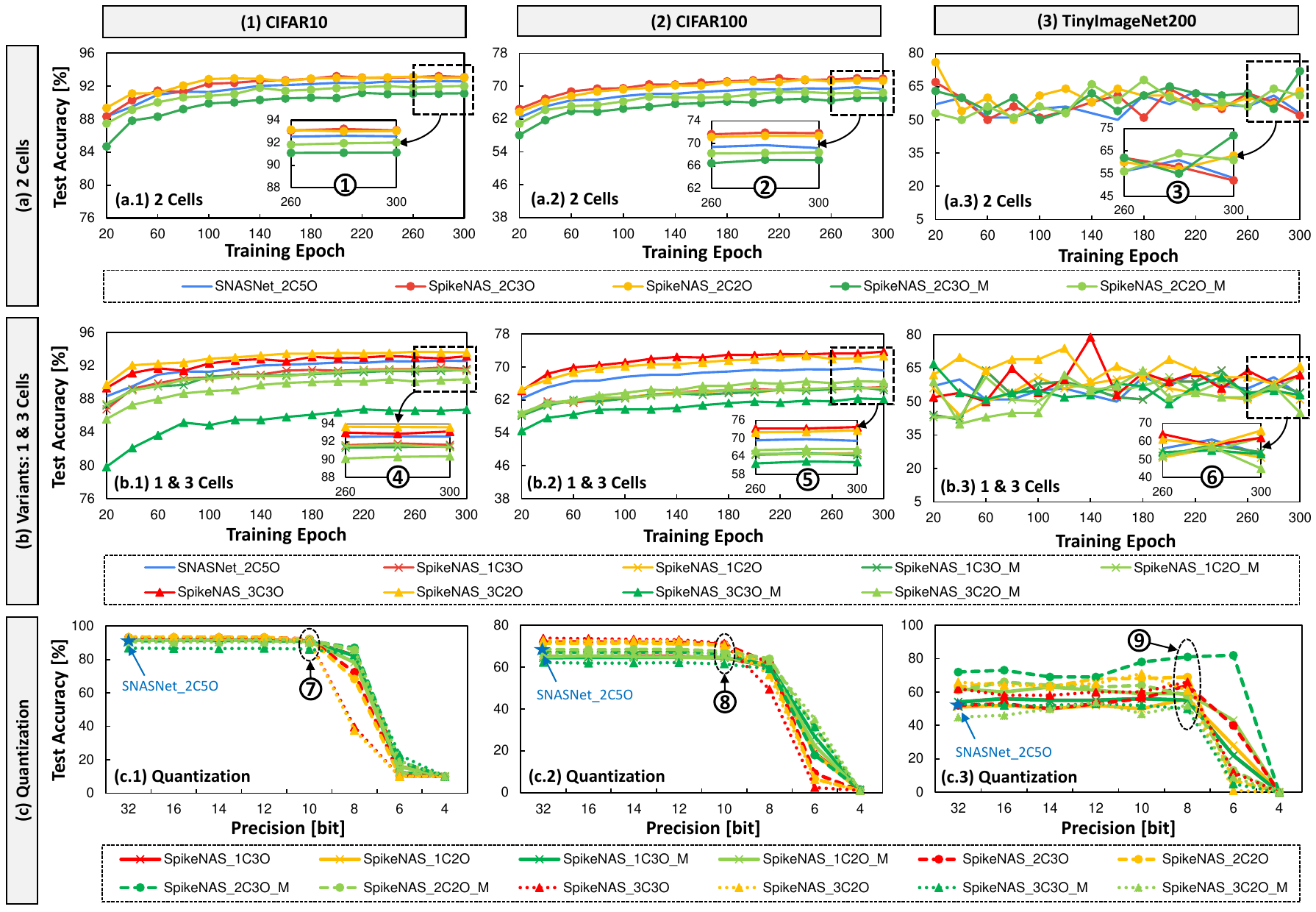}
\vspace{-0.7cm}
\caption{Results of the accuracy for SNNs with (a) 2 cells, (b) variants of 1 \& 3 cells, and (c) quantized weights across different NAS techniques, settings, and datasets: CIFAR10, CIFAR100, and TinyImageNet200.}
\label{Fig_Results_Acc}
\vspace{-0.6cm}
\end{figure*}

For evaluation and validation of the proposed concepts, we develop a Python-based implementation using the SpikingJelly library~\cite{Ref_Fang_SpikingJelly_SciAdv23} on the Nvidia RTX A6000 GPU machine with 48GB GDDR6 memory. 
We consider the CIFAR10, CIFAR100, and TinyImageNet200 datasets as workloads which represent simple, medium, and complex tasks, respectively.
We perform up to 300 training epochs with the surrogate gradient learning~\cite{Ref_Neftci_SurrogateSNNs_IEEEMSP19}, while evaluating the test accuracy gradually to see the learning curve of the investigated network architecture. 
These experiments provide information of accuracy, searching time, memory footprint, and power consumption. 
These searching time and power are then leveraged to estimate the energy consumption. 
Meanwhile, for evaluating the proposed concepts on embedded platforms, we employ the Python-based implementation on the Nvidia Jetson AGX Orin.

\vspace{-0.2cm}
\section{Results and Discussion}
\label{Sec_Results}

\subsection{Maintaining High Accuracy}
\label{Sec_Results_Accuracy}

Fig.~\ref{Fig_Results_Acc}(a)-(b) present the accuracy for SNNs with 2 cells and variants (i.e., 1 and 3 cells), respectively.
Meanwhile, Fig.~\ref{Fig_Results_Acc}(c) presents the accuracy for quantized SNNs.

\smallskip
\textbf{The 2-Cell Architectures:} 
SNASNet achieves 92.58\% of accuracy for CIFAR10, 69.18\% for CIFAR100, and 53\% for TinyImageNet200.  
Our SpikeNAS\_2C3O achieves 93.10\% of accuracy for CIFAR10, 71.81\% for CIFAR100, and 52\% for TinyImageNet200;   
while SpikeNAS\_2C2O achieves 93.05\% of accuracy for CIFAR10, 71.34\% for CIFAR100, and 63\% for TinyImageNet200; see \circled{1}, \circled{2}, and \circled{3}.
These results indicate that, our SpikeNAS maintains (or improves in most cases) the accuracy as compared to the state-of-the-art, since SpikeNAS only considers operations with high significance to the accuracy.  
Moreover, these improvements can be achieved by networks from SpikeNAS with less than 10M parameters.
When the memory constraints are given, our SpikeNAS\_2C3O\_M achieves 91.13\% of accuracy for CIFAR10, 67.06\% for CIFAR100, and 72\% for TinyImageNet200; 
while SpikeNAS\_2C2O\_M achieves 92.02\% of accuracy for CIFAR10, 68.41\% for CIFAR100, and 61\% for TinyImageNet200; see \circled{1}, \circled{2}, and \circled{3}.
These results show that, our SpikeNAS maintains (or improves in some cases) the accuracy as compared to the state-of-the-art, while ensuring that the generated SNN architecture meets the memory budget. 
The reason is that, SpikeNAS employs operations with high significance to the accuracy and incorporates the memory budget in its search process.
Note, memory evaluation is discussed in Section~\ref{Sec_Results_Memory}.

\begin{figure*}[t]
\centering
\includegraphics[width=\linewidth]{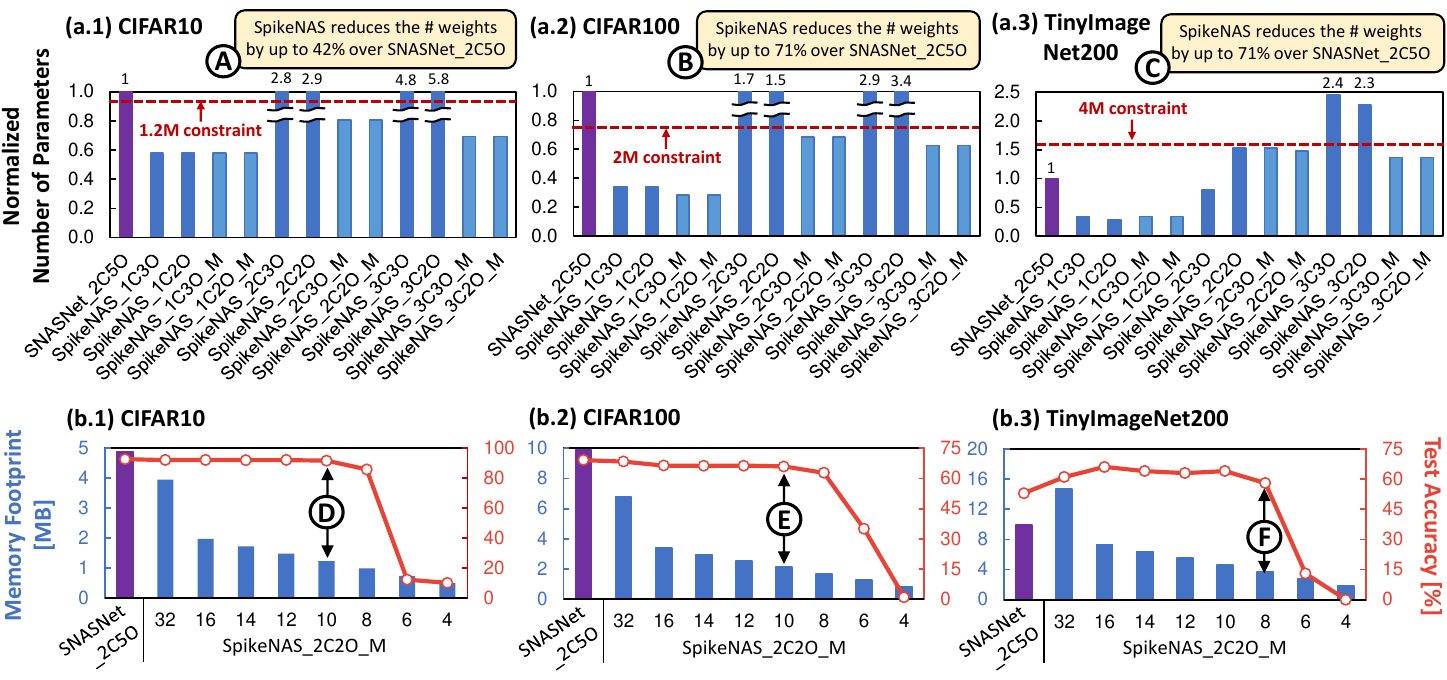}
\vspace{-0.8cm}
\caption{(a) Results of the number of parameters normalized to the state-of-the-art work SNASNet\_2C5O across different NAS techniques, settings, and datasets. (b) Accuracy-memory trade-offs between state-of-the-art work SNASNet\_2C5O with our SpikeNAS\_2C2O\_M under different precision levels and across different datasets: CIFAR10, CIFAR100, and TinyImageNet200.}
\label{Fig_Results_Mem}
\vspace{-0.6cm}
\end{figure*}

\smallskip
\textbf{The 1-Cell and 3-Cell Architectures (Variants):} 
Our SpikeNAS\_1C3O achieves 91.65\% of accuracy for CIFAR10, 65.10\% for CIFAR100, and 54\% for TinyImageNet200; 
while SpikeNAS\_1C2O achieves 91.41\% of accuracy for CIFAR10, 64.92\% for CIFAR100, and 51\% for TinyImageNet200; see \circled{4}, \circled{5}, and \circled{6}. 
Furthermore, our SpikeNAS\_3C3O achieves 93.16\% of accuracy for CIFAR10, 73.74\% for CIFAR100, and 62\% for TinyImageNet200; 
while SpikeNAS\_3C2O achieves 93.63\% of accuracy for CIFAR10, 72.65\% for CIFAR100, and 66\% for TinyImageNet200; see \circled{4}, \circled{5}, and \circled{6}. 
These results indicate that, in general, the 1-cell architectures offer a slightly lower accuracy as compared to the 2-cell architectures, as the 1-cell architecture have fewer operations to extract meaningful features from the inputs. 
Meanwhile, 3-cell architectures can offer a higher accuracy as compared to the 2-cell architectures, as the 3-cell architecture have more operations to extract meaningful features from the inputs.
When the memory constraints are given, SpikeNAS\_1C3O\_M and SpikeNAS\_1C2O\_M have similar results to their unconstrained counterparts (i.e., SpikeNAS\_1C3O and SpikeNAS\_1C2O) for both CIFAR10 and CIFAR100, since the memory footprints of the unconstrained ones are already within the given memory budgets, and hence leading to similar network architectures and accuracy.
Meanwhile, SpikeNAS\_3C3O\_M and SpikeNAS\_3C2O\_M have lower accuracy than their unconstrained counterparts (i.e., SpikeNAS\_3C3O and SpikeNAS\_3C2O), since the memory constraint adds a limitation for finding a combination of cell architectures that offers high accuracy. 

\smallskip
\textbf{Accuracy under Quantization:}
In general, applying lower bit precision levels leads to accuracy degradation due to the reduced capabilities of representing more learned knowledge (information) in the weights.
Here, we observe that employing at least 10-bit precision for CIFAR10 \circled{7} and CIFAR100 \circled{8} as well as at least 8-bit precision for TinyImageNet200 \circled{9} maintains the accuracy comparable to that of the state-of-the-art (and even improves the accuracy in some cases), indicating that these precision levels are sufficient for preserving the learned knowledge (i.e., input features).

\smallskip
These results highlight that, \textit{our SpikeNAS can generate SNN architectures that maintain or improve the accuracy as compared to the state-of-the-art}. 

\vspace{-0.3cm}
\subsection{Ensuring Memory-aware SNN Generation}
\label{Sec_Results_Memory}

Fig.~\ref{Fig_Results_Mem}(a) presents the experimental results for the number of parameters across different NAS techniques, settings, and datasets.
For CIFAR10, SNASNet incurs 6\% memory overhead than the given budget, while SpikeNAS can save the memory footprint up to 42\% from the state-of-the-art; see \circled{A}. 
If we consider high accuracy, then we can select SpikeNAS\_2C2O\_M with 92.02\% accuracy and 19\% memory saving from the state-of-the-art (i.e., 14\% memory saving from the given budget).
For CIFAR100, SNASNet incurs 23\% memory overhead than the given budget, while SpikeNAS can save the memory footprint up to 71\% from the state-of-the-art; see \circled{B}. 
If we consider high accuracy, then we can select SpikeNAS\_2C2O\_M with 68.41\% accuracy and 32\% memory saving from the state-of-the-art (11\% memory saving from the given budget).
Meanwhile, for TinyImageNet200, SNASNet and our SpikeNAS meet the given memory budget. 
Moreover, SpikeNAS can save the memory footprint up to 71\% from the state-of-the-art; see \circled{C}. 
If we consider high accuracy, then we can select SpikeNAS\_2C3O\_M with 72\% accuracy and 0.4\% memory saving from the given budget.

Furthermore, our quantization further reduces the memory footprint of SNN models. 
Fig.~\ref{Fig_Results_Mem}(b) presents the results of experimental case study that compares SNASNet and our SpikeNAS\_2C2O\_M under different precision levels.
Here, we observe that the high accuracy of SpikeNAS\_2C2O\_M can be maintained as long as at least 10-bit precision is employed for CIFAR10 and CIFAR100, hence leading to 75\% and 79\% memory savings over SNASNet, respectively.  
Meanwhile for TinyImageNet200, the high accuracy of SpikeNAS\_2C2O\_M can be maintained as long as at least 8-bit precision is employed, hence leading to 63\% memory saving over SNASNet.

These results highlight that, \textit{SpikeNAS can generate SNN architectures that meet the given memory constraints due to its memory-aware search algorithm}, enabling efficient deployments of SNN models on the target hardware platforms. 

\vspace{-0.3cm}
\subsection{Comparing Accuracy and Memory of NAS for SNNs}
\label{Sec_Results_TradeOff}

To show the advancements of NAS for SNNs in terms of accuracy and memory requirements, we compare our SpikeNAS against different state-of-the-art works (i.e., AutoSNN~\cite{Ref_Na_AutoSNN_ICML22}, AutoSpikformer~\cite{Ref_Che_AutoSpikeformer_arXiv23}, AutoST~\cite{Ref_Wang_AutoST_ICASSP24}, and SNASNet~\cite{Ref_Kim_SNASNet_ECCV22}); see a summary in Table~\ref{Tab_Results}. 
If we consider memory constraints of 1.2M parameters for CIFAR10, 2M parameters for CIFAR100, and 4M parameters for TinyImageNet200, then our SpikeNAS achieves competitive accuracy as compared to other NAS methods, while meeting the given memory constraints. 
Meanwhile, other NAS methods achieve high accuracy but at the cost of significantly larger memory requirements (i.e., higher number of parameters) than our SpikeNAS, hence they fail to meet the given memory constraints. 
The reason is that, only our SpikeNAS framework has the focus on hardware resources in the NAS process, hence having the capabilities to meet the given memory constraints.

\begin{table}[hbtp]
\vspace{-0.2cm}
\caption{Comparison of different state-of-the-art NAS methods for SNNs on CIFAR10, CIFAR100, and TinyImageNet200. \\
Note: results for~\cite{Ref_Na_AutoSNN_ICML22, Ref_Che_AutoSpikeformer_arXiv23, Ref_Wang_AutoST_ICASSP24} are obtained from the respective papers; results for~\cite{Ref_Kim_SNASNet_ECCV22} and our SpikeNAS are obtained from our experiments;
N/A means no available information; while results from hardware-aware NAS are written in \textbf{bold}. (*) Only our SpikeNAS considers hardware resources in the NAS process.}
\vspace{-0.1cm}
\centering
\label{Tab_Results}
\footnotesize
\begin{tabular}{|c|cc|cc|cc|}
\hline
\multicolumn{1}{|c|}{\multirow{2}{*}{Work}} & \multicolumn{2}{c|}{CIFAR10} & \multicolumn{2}{c|}{CIFAR100} & \multicolumn{2}{c|}{TinyImageNet200} \\ 
\cline{2-7} 
\multicolumn{1}{|c|}{} & \multicolumn{1}{c|}{\begin{tabular}[c]{@{}c@{}}\# Param.\\ {[}M{]}\end{tabular}} & \multicolumn{1}{c|}{\begin{tabular}[c]{@{}c@{}}Acc.\\ {[}\%{]}\end{tabular}} & \multicolumn{1}{c|}{\begin{tabular}[c]{@{}c@{}}\# Param.\\ {[}M{]}\end{tabular}} & \multicolumn{1}{c|}{\begin{tabular}[c]{@{}c@{}}Acc.\\ {[}\%{]}\end{tabular}} & \multicolumn{1}{c|}{\begin{tabular}[c]{@{}c@{}}\# Param.\\ {[}M{]}\end{tabular}} & \multicolumn{1}{c|}{\begin{tabular}[c]{@{}c@{}}Acc.\\ {[}\%{]}\end{tabular}} \\ 
\hline
\hline
\cite{Ref_Na_AutoSNN_ICML22} & \begin{tabular}[c]{@{}c@{}}1.46\\ 20.92\end{tabular} & \begin{tabular}[c]{@{}c@{}}91.32\\ 93.15\end{tabular} & N/A & 69.16 & N/A & 46.79 \\
\hline
\cite{Ref_Che_AutoSpikeformer_arXiv23} & \begin{tabular}[c]{@{}c@{}}4.69\\ 9.20\end{tabular} & \begin{tabular}[c]{@{}c@{}}95.29\\ 96.38\end{tabular} & \begin{tabular}[c]{@{}c@{}}4.69\\ 9.20\end{tabular} & \begin{tabular}[c]{@{}c@{}}77.71\\ 77.05\end{tabular} & N/A & N/A \\
\hline
\cite{Ref_Wang_AutoST_ICASSP24} & \begin{tabular}[c]{@{}c@{}}11.52\\ 29.64\end{tabular} & \begin{tabular}[c]{@{}c@{}}96.03\\ 96.21\end{tabular} & \begin{tabular}[c]{@{}c@{}}11.52\\ 29.64\end{tabular} & \begin{tabular}[c]{@{}c@{}}79.44\\ 79.69\end{tabular} & \begin{tabular}[c]{@{}c@{}}5.99\\ 34.44\end{tabular} & \begin{tabular}[c]{@{}c@{}}63.80\\ 74.54\end{tabular} \\
\hline
\cite{Ref_Kim_SNASNet_ECCV22} & 1.28 & 92.58 & 2.59 & 69.18 & 2.59 & 53.00 \\
\hline
Ours$^{*}$ & \begin{tabular}[c]{@{}c@{}}\textbf{1.04}\\ 7.38\end{tabular} & \begin{tabular}[c]{@{}c@{}}\textbf{92.02}\\ 93.63\end{tabular} & \begin{tabular}[c]{@{}c@{}}\textbf{1.77}\\ 7.67\end{tabular} & \begin{tabular}[c]{@{}c@{}}\textbf{68.41}\\ 73.74\end{tabular} & \begin{tabular}[c]{@{}c@{}} \textbf{3.98}\\ 5.90\end{tabular} & \begin{tabular}[c]{@{}c@{}} \textbf{72.00}\\ 66.00\end{tabular} \\
\hline
\end{tabular}
\vspace{-0.2cm}
\end{table}

\vspace{-0.3cm}
\subsection{Reducing the Searching Time}
\label{Sec_Results_SearchTime}

\begin{figure*}[t]
\centering
\includegraphics[width=\linewidth]{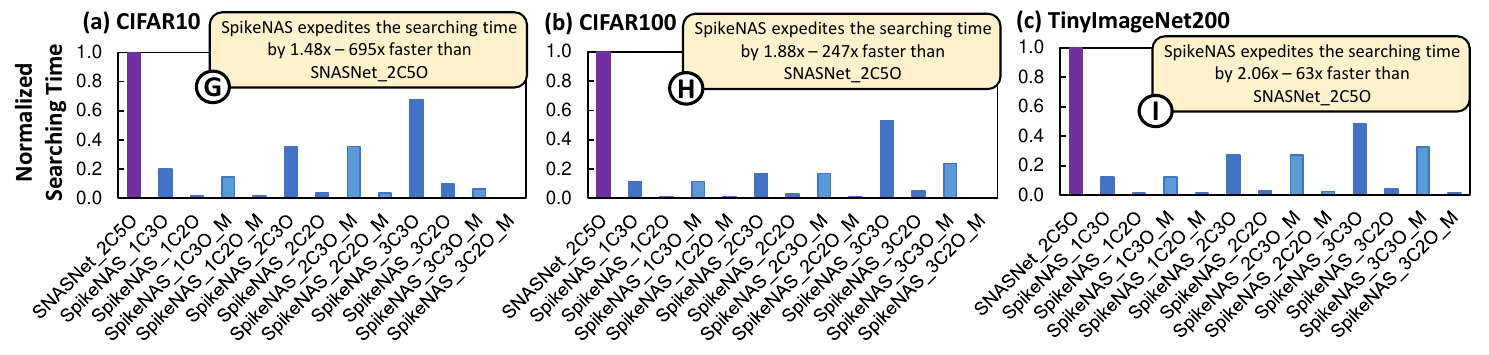}
\vspace{-0.6cm}
\caption{Results of the searching time normalized to the state-of-the-art work SNASNet\_2C5O for (a) CIFAR10, (b) CIFAR100, and (c) TinyImageNet200 across different NAS techniques and settings.}
\label{Fig_Results_Time}
\vspace{-0.2cm}
\end{figure*}

Fig.~\ref{Fig_Results_Time} presents the experimental results of the searching time.
In all cases of workloads (i.e., CIFAR10, CIFAR100, and TinyImageNet200), our SpikeNAS offers significant improvements for the searching time as compared to the state-of-the-art SNASNet.
Specifically, our SpikeNAS expedites the searching time by 1.48x-695x for CIFAR10 \circled{G}, by 1.88x-247x for CIFAR100 \circled{H}, and by 2.06x-63x for TinyImageNet200 \circled{I}.
The reason is that, SpikeNAS employs fewer operation types that have significant impact on the accuracy, performs a consecutive search across the existing cells, and incorporates a memory constraint into the search process, thereby leading to a smaller search space and a faster searching time than the state-of-the-art.
If we consider high accuracy, acceptable memory cost, and fast searching time, then we can select SpikeNAS\_2C2O\_M for CIFAR10 with 92.02\% accuracy, 14\% memory saving from the given memory budget, and 29x faster search than the state-of-the-art. 
For CIFAR100, we can select SpikeNAS\_2C2O\_M with 68.41\% accuracy, 11\% memory saving from the given memory budget, and 117x faster search than the state-of-the-art.
Meanwhile for TinyImageNet200, we can select SpikeNAS\_2C3O\_M with 72\% accuracy, 0.38\% memory saving from the given memory budget, and 3.7x faster search than the state-of-the-art.

These results highlight that, \textit{our SpikeNAS can quickly generate SNN architectures that offer high accuracy and meet the memory constraints}, hence enabling efficient design automation for developing SNN-based embedded AI systems. 

\begin{figure*}[t]
\centering
\includegraphics[width=\linewidth]{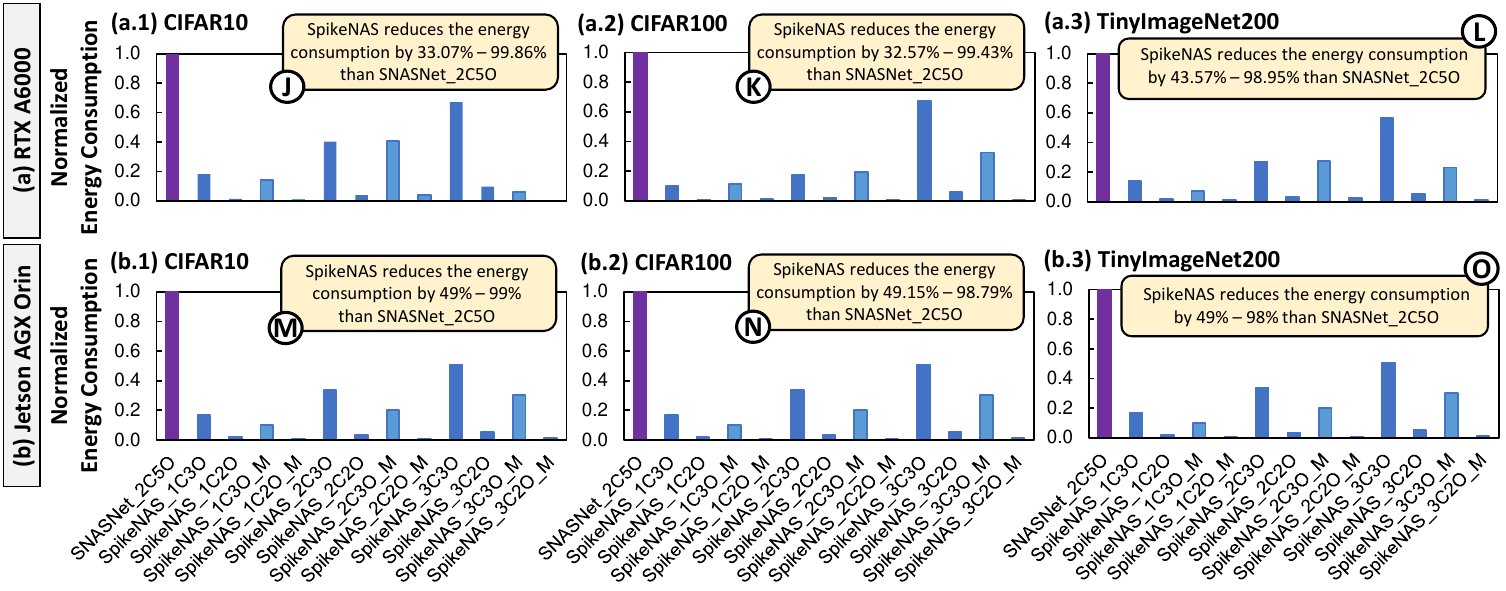}
\vspace{-0.6cm}
\caption{Results of the energy consumption normalized to the state-of-the-art work SNASNet\_2C5O for SNN implementation on (a) Nvidia RTX A6000 GPU machine, and (b) Nvidia Jetson AGX Orin, across different NAS techniques, settings, and datasets:  CIFAR10, CIFAR100, and TinyImageNet200.}
\label{Fig_Results_Energy}
\vspace{-0.5cm}
\end{figure*}

\vspace{-0.2cm}
\subsection{Improving Energy Efficiency}
\label{Sec_Results_Efficiency}

Fig.~\ref{Fig_Results_Energy} presents the experimental results of the energy consumption for SNN implementation on Nvidia RTX A6000 GPU machine (represents a desktop platform) and Nvidia Jetson AGX Orin (represents an embedded platform).
In general, our SpikeNAS incurs significantly lower energy consumption as compared to the state-of-the-art SNASNet in both desktop and embedded platforms, therefore leading to notable energy efficiency improvements. 
Specifically, in the RTX A6000 GPU machine, our SpikeNAS reduces energy consumption by 33.07\%-99.86\% (i.e., 1.49x-711x energy efficiency improvements) for CIFAR10 \circled{J}, by 32.57\%-99.43\% (i.e., 1.48x-17x energy efficiency improvements) for CIFAR100 \circled{K}, and by 43.57\%-98.95\%x (i.e., 1.77x-95x energy efficiency improvements) for TinyImageNet200 \circled{L}. 
Meanwhile, in the Jetson AGX Orin, our SpikeNAS reduces energy consumption by about 49\%-99\% (i.e., 2x-82x energy efficiency improvements) for CIFAR10 \circled{M}, CIFAR100 \circled{N}, and TinyImageNet200 \circled{O}. 
These energy savings come from the effectiveness of our proposed optimization techniques in SpikeNAS that limit the network size based on the memory constraint, which lead to the reduction of processing power and processing time, and hence improving the energy efficiency. 

\vspace{-0.3cm}
\subsection{Flexibility of SpikeNAS Framework}
\label{Sec_Results_Discussion}
    
In the SNN community, many different optimization techniques have been explored to improve the energy efficiency of SNN processing~\cite{Ref_Yin_SATA_TCAD22, Ref_Yin_MINT_ASPDAC24, {Ref_Yin_PruningSNNs_TETCI24}}, and they also have the potential to be incorporated into our SpikeNAS framework to further enhance the efficiency of SNN models. 
For instance, the SATA accelerator~\cite{Ref_Yin_SATA_TCAD22} can be employed to expedite the training of SNN architectures that are generated from SpikeNAS, since both SATA accelerator and SpikeNAS framework consider the same surrogate gradient learning~\cite{Ref_Neftci_SurrogateSNNs_IEEEMSP19} for SNN training. 
The MINT method~\cite{Ref_Yin_MINT_ASPDAC24} can be employed in the quantization step of SpikeNAS for providing more options of quantization modes and memory saving, since MINT mainly focuses on quantizing the membrane potentials of LIF neurons.
Meanwhile, a workload-balanced pruning technique~\cite{Ref_Yin_PruningSNNs_TETCI24} can be performed after the training phase in SpikeNAS to further compress the SNN model while ensuring a balanced-workload during the inference phase. 
Therefore, SpikeNAS framework has high flexibility for integration with other optimization techniques. 

Our proposed techniques in the SpikeNAS framework (e.g., reduction of network operations) could be applicable to ANNs. 
For instance, the work of~\cite{Ref_Yang_AlwaysOnDNN_JSSC19} may reduce the number of operations through pruning to further improve its efficiency. 
However, SNN-based solutions can offer alternate and unique advantages over ANN-based solutions, such as their ultra-low power/energy computation due to sparse spike-based computation and their inherent spatio-temporal learning capabilities, which are especially important for resource-constrained embedded AI systems~\cite{Ref_Minhas_NCLsurvey_arXiv24}.
To enable the SNN applicability for diverse application use-cases, SNN developments should be performed considering the design requirements (e.g., memory constraint), and supported by the bio-inspired computing platform with minimum/no power-hungry hardware like Analog-to-Digital Converters (ADCs)~\cite{Ref_Minhas_NCLsurvey_arXiv24}\cite{Ref_Putra_SNNonCNP_arXiv25}.
Therefore, our SpikeNAS framework enables a fast memory-aware NAS in developing SNN model that meets the given design requirements.
Since NAS techniques for SNNs have not been extensively explored, our SpikeNAS framework may pave a way in enabling efficient SNN design targeting resource-constrained embedded AI systems. 

\section{Conclusion}
\label{Sec_Conclusion}

We propose the SpikeNAS framework to quickly find an SNN architecture that provides high accuracy under a memory constraint from the hardware platform of the embedded AI system. 
SpikeNAS employs an analysis of the impact of operations on the accuracy, architecture enhancements, and a fast search algorithm that is guided by memory budget and consecutive searches across cells. 
The experimental results show that, SpikeNAS maintains the accuracy and improves the searching time as compared to the state-of-the-art with a smaller memory cost. 
As a result, our SpikeNAS improves the searching time and maintains high accuracy compared to state-of-the-art while meeting the given memory budgets (e.g., 29x, 117x, and 3.7x faster search for CIFAR10, CIFAR100, and TinyImageNet200, respectively).
In this manner, our SpikeNAS framework enables the efficient SNN deployments in embedded AI systems for diverse application use-cases.


% ...

\section*{Acknowledgment}
\label{Sec_Ack}

This work was partially supported by the NYUAD Center for Artificial Intelligence and Robotics (CAIR), funded by Tamkeen under the NYUAD Research Institute Award CG010.


\bibliographystyle{IEEEtran}
\bibliography{bibliography}


\begin{IEEEbiography}[{\includegraphics[width=1in,height=1.25in,clip,keepaspectratio]{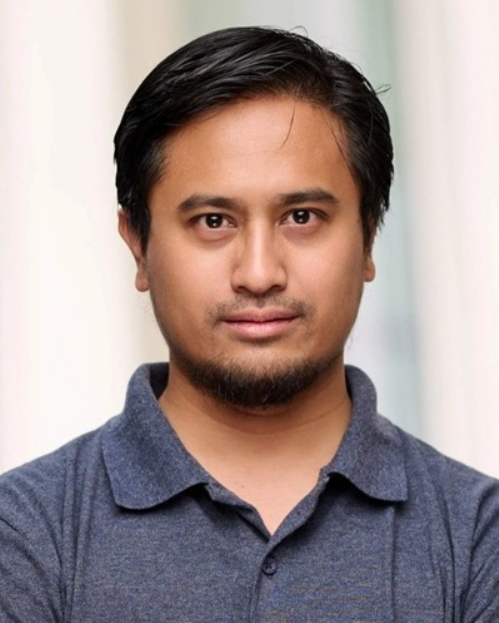}}]{Rachmad Vidya Wicaksana Putra}
(Member, IEEE)
received Ph.D. degree in Computer Science from Technische Universit\"at Wien (TU Wien), Vienna, Austria.
Previously, he received B.Sc. degree in Electrical Engineering and M.Sc. degree in Electronics Engineering with a Cum Laude distinction from Bandung Institute of Technology (ITB), Bandung, Indonesia. 
In academia, he experienced working as a teaching assistant at the Electrical Engineering Department ITB, a research assistant at the Microelectronics Center ITB, and a project research assistant at the Institute of Computer Engineering TU Wien. 
Meanwhile in industry, he experienced working as an FPGA engineer at companies in Indonesia and Austria. 
Currently, he is a Research Team Leader at the eBrain Lab, New York University (NYU) Abu Dhabi, Abu Dhabi, UAE. 
He served as reviewer in several highly reputed journals from IEEE, ACM, and Springer, as well as prestigious IEEE conferences. 
He received multiple HiPEAC Paper Awards and an ACM Showcase for his first-authored papers during his Ph.D. study, and a best paper nomination in ICARCV 2024. 
His research interests include brain-inspired computing, neuroAI, neuromorphic engineering, computer architectures, integrated circuits \& VLSI, system-on-chip, emerging device technologies, hardware/software co-design for robust \& energy-efficient systems, and embedded AI.
\end{IEEEbiography}

\begin{IEEEbiography}[{\includegraphics[width=1in,height=1.25in,clip,keepaspectratio]{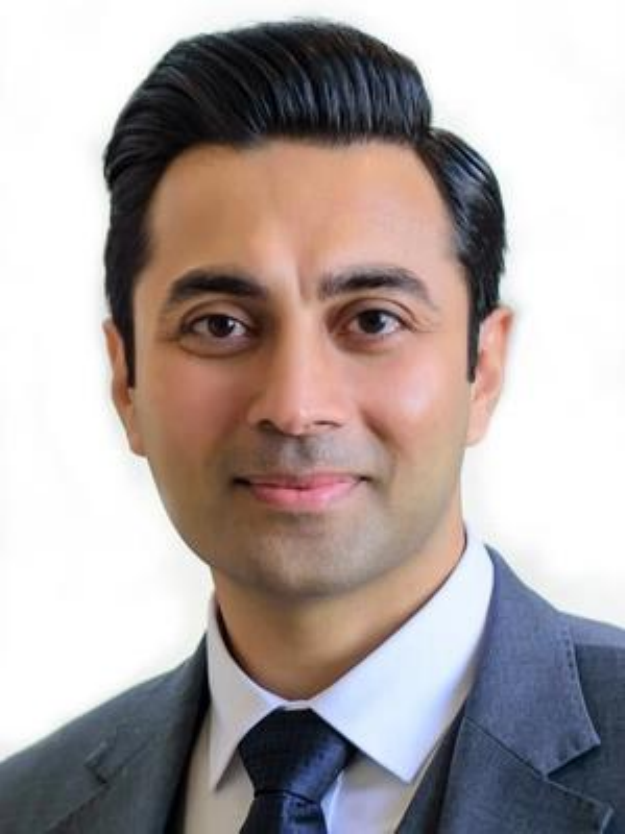}}]{Muhammad Shafique}
(Senior Member, IEEE)
received the Ph.D. degree in computer science from
Karlsruhe Institute of Technology (KIT), Karlsruhe,
Germany, in 2011.

Afterwards, he established and led a highly recognized research group at KIT for several years as well as conducted impactful collaborative R\&D activities across the globe. 
In October 2016, he joined as a Full Professor in computer architecture and robust, energy-efficient technologies at the Faculty of Informatics,
Institute of Computer Engineering, Technische Universit\"at Wien (TU Wien), Vienna, Austria. 
Since September 2020, he has been with New York University (NYU) Abu Dhabi, Abu Dhabi, UAE, where he is currently a Full Professor and the Director of eBrain Lab. 
He is a Global Network Professor with Tandon School of Engineering, NYU, New York, NY, USA. 
He is also a Co-PI/Investigator with multiple NYUAD Centers, including Center of Artificial Intelligence and Robotics (CAIR), Center of Cyber Security (CCS), Center for InTeractIng urban nEtworkS (CITIES), and Center for Quantum and Topological Systems (CQTS). 
His research interests include AI \& machine learning hardware and system-level design, brain-inspired computing, quantum machine learning, cognitive autonomous systems, wearable healthcare, energy-efficient systems, robust computing, hardware security, emerging technologies, FPGAs, MPSoCs, and embedded systems. 
His research has a special focus on cross-layer analysis, modeling, design, and optimization of computing and memory systems. 
The researched technologies and tools are deployed in application use cases from Internet-of-Things (IoT), smart cyber–physical systems (CPSs), and ICT for development (ICT4D) domains. 
He has given several keynotes, invited talks, and tutorials, as well as organized many special sessions at premier venues. 
He has served as the PC Chair, the General Chair, the Track Chair, and a PC member for several prestigious IEEE/ACM conferences. 
He holds one U.S. patent and has (co-)authored 10 books, 25+ book chapters, 400 papers in premier journals and conferences, and 170+ archive articles.

Dr. Shafique received the 2015 ACM/SIGDA Outstanding New Faculty
Award, the AI 2000 Chip Technology Most Influential Scholar Award in
2020, 2022, and 2023, the ASPIRE AARE Research Excellence Award in 2021,
six gold medals, and several best paper awards and nominations at prestigious conferences. 
He is a senior member of the IEEE Signal Processing Society (SPS) and a member of the ACM, SIGARCH, SIGDA, SIGBED, and HIPEAC.
\end{IEEEbiography}

\end{document}